\documentclass[twoside]{article}
\usepackage[preprint]{aistats2024}
\usepackage[round]{natbib}

\usepackage[dvipsnames]{xcolor} 

\usepackage{xcolor}

\newcommand{\norm}[1]{\left\Vert#1\right\Vert}

\newcommand{\parr}[1]{\left (#1\right )}

\newcommand{\Real}{\mathbb R}

\newcommand{\eps}{\varepsilon}
\newcommand{\too}{\rightarrow}

\definecolor{mygray}{gray}{0.95}

\newcommand{\ie}{{i.e.}}

\usepackage{amsmath,amsfonts,bm}

\let\oldeqref\eqref
\def\eqref#1{equation~\oldeqref{#1}}

\def\1{\bm{1}}

\def\eps{{\epsilon}}

\def\vs{{\bm{s}}}

\DeclareMathAlphabet{\mathsfit}{\encodingdefault}{\sfdefault}{m}{sl}
\SetMathAlphabet{\mathsfit}{bold}{\encodingdefault}{\sfdefault}{bx}{n}

\def\gL{{\mathcal{L}}}

\def\gN{{\mathcal{N}}}

\newcommand{\E}{\mathbb{E}}

\newcommand{\R}{\mathbb{R}}

\usepackage{tabularx,ragged2e,booktabs,caption}
\newcolumntype{C}[1]{>{\Centering}m{#1}}

\newcolumntype{Z}[1]{>{\Left}m{#1}}

\usepackage[colorlinks=true,citecolor=Blue, pageanchor=true,plainpages=false,pdfpagelabels,bookmarks,bookmarksnumbered]{hyperref}
\usepackage{url}
\usepackage{cleveref}
\usepackage{color, float}
\usepackage[ruled, vlined]{algorithm2e}
\usepackage{multirow}
\usepackage{tcolorbox}
\usepackage{graphicx}
\usepackage{enumitem}
\usepackage{amsmath, amsfonts, amssymb, mathtools, dsfont}
\usepackage{thmtools,thm-restate}
\usepackage{bm}
\usepackage{wrapfig}
\usepackage[font=small,labelfont=bf]{caption}
\usepackage{arydshln}
\usepackage{subcaption}
\usepackage{tabu}
\usepackage{footnote}
\usepackage{algpseudocode}
\graphicspath{{./figure/}}
\usepackage[dvipsnames]{xcolor}

\newcommand{\ahat}{\widehat{a}}
\newcommand{\shat}{\widehat{s}}
\renewcommand{\hat}{\widehat}

\newcommand{\hopper}{\texttt{hopper}\xspace}
\newcommand{\walker}{\texttt{walker}\xspace}
\newcommand{\cheetah}{\texttt{halfcheetah}\xspace}

\newcommand{\gym}{\texttt{Gym}\xspace}
\newcommand{\medexpert}{\texttt{medium-expert}\xspace}
\newcommand{\medreplay}{\texttt{medium-replay}\xspace}
\newcommand{\medium}{\texttt{medium}\xspace}
\newcommand{\med}{\texttt{medium}\xspace}

\newcommand{\N}{\mathcal{N}}
\newcommand{\A}{\mathcal{A}}

\renewcommand{\S}{\mathcal{S}}

\renewcommand{\vs}{\bm{s}}

\newcommand{\node}{n_\text{ode}}
\renewcommand{\cite}{\citep}
\definecolor{lightorange}{HTML}{ff7f2a}
\definecolor{lighterorange}{HTML}{ffe6d5}
\newtcolorbox{summarybox}{colback=lighterorange,colframe=lightorange,
boxsep=1pt,left=3pt,right=3pt,top=3pt,bottom=3pt}
\newcommand{\highlight}[1]{\colorbox{Goldenrod!30}{\textbf{#1}}}
\newcommand{\tausub}{\tau_\text{sub}}

\newcommand{\qtilde}{\tilde{q}}
\newcommand{\ptilde}{\tilde{p}}
\newcommand{\ut}{u^\theta}

\counterwithin{figure}{section}
\counterwithin{table}{section}
\counterwithin{algocf}{section}

\usepackage{titlesec}
\titlespacing{\section}{0pt}{5pt}{0pt}
\titlespacing{\subsection}{0pt}{5pt}{0pt}
\titlespacing{\subsubsection}{0pt}{5pt}{0pt}
\titlespacing{\paragraph}{0pt}{0pt}{5pt}
\usepackage{authblk}

\renewcommand*{\Affilfont}{\normalsize}
\makeatletter
\renewcommand\AB@affilsepx{\quad \protect\Affilfont}
\makeatother

\author[1]{{\normalsize  Qinqing Zheng}}
\author[1]{{\normalsize  Matt Le}}
\author[2]{{\normalsize  Neta Shaul}}
\author[1,2]{{\normalsize  Yaron Lipman}}
\author[3]{{\normalsize Aditya Grover}}
\author[1]{{\normalsize Ricky T. Q. Chen}}

\affil[1]{FAIR, Meta}
\affil[2]{Weizmann Institute of Science}
\affil[3]{UCLA}
\title{\vspace{-50pt}}
\date{}
\begin{document}

\twocolumn[
\aistatstitle{Guided Flows for Generative Modeling and Decision Making}
\maketitle
]

\begin{abstract}
Classifier-free guidance is a key component for enhancing the performance of conditional generative models across diverse tasks. While it has previously demonstrated remarkable
improvements for the sample quality, it has only been exclusively employed for diffusion models. In this paper, we integrate classifier-free guidance into Flow Matching (FM) models, an alternative simulation-free approach that trains Continuous Normalizing Flows (CNFs) based on regressing vector fields. We explore the usage of \emph{Guided Flows} for a variety of downstream applications. We show that Guided Flows significantly improves the sample quality in conditional image generation and zero-shot text-to-speech synthesis, boasting state-of-the-art performance. Notably, we are the first to apply flow models for plan generation in the offline reinforcement learning setting, showcasing a 10x speedup in computation compared to diffusion models while maintaining comparable performance.
\end{abstract}
\section{Introduction}
\label{sec:intro}

Conditional generative modeling paves the way to numerous machine learning applications such as conditional image generation~\citep{dhariwal2021diffusion,rombach2022high}, text-to-speech synthesis~\citep{wang2023neural,le2023voicebox}, and even solving decision making problems \citep{chen2021decision, janner2021offline, janner2022planning, ajay2022conditional}. Models that appear ubiquitously across a variety of application domains are diffusion models \citep{sohl2015deep,ho2020denoising} and flow-based models \citep{song2020score,lipman2022flow,albergo2022building}. Majority of this development has been focused around diffusion models, where multiple forms of conditional guidance~\cite{dhariwal2021diffusion,ho2022classifier} have been introduced to place larger emphasis on the conditional information.
While flow models have been shown to be more efficient alternatives than diffusion models~\citep{lipman2022flow,pooladian2023multisample} in unconditional generation, requiring less computation to sample, their behavior in conditional generation tasks has not been explored as much. It also remains unclear whether conditional guidance can be applied to and help the performance of flow-based models.


In this work, we study the behavior of Flow Matching models for conditional generation. We introduce \emph{Guided Flows}, an adaptation of classifier-free guidance \citep{ho2022classifier} to Flow Matching models, showing that an analogous modification can be made to the velocity vector fields, including the optimal transport~\citep{lipman2022flow} and cosine scheduling~\citep{albergo2022building} flows used by prior works.

\begin{table}[t]
    \centering
    \resizebox{\columnwidth}{!}{
    \begin{tabular}{c c c}
    \toprule
    Application     &  data point $x_1$  & conditioning variables $y$ \\
    \midrule
    Image Generation & image & class label \\
    Text-to-Speech & spectrogram & text \& utterance \\
    Offline RL & state sequence & target return \\
    \bottomrule
    \end{tabular}
    }
    \caption{The data point $x_1$ and the conditioning variables $y$ on which the conditional guidance will be applied, for the three applications settings that we consider.
    }
    \label{tab:x_y_by_application}
\end{table}

We experimentally validate Guided Flows on a variety of applications, ranging from generative modeling over multiple modalities to offline reinforcement learning (RL), see Table~\ref{tab:x_y_by_application}. As we show in Section~\ref{sec:image}, for standard generative tasks including image synthesis and zero-shot text-to-speech generation, Guided Flows significantly improves the sampling quality over unguided counterparts (i.e., sample from the conditional distribution directly), attaining state-of-the-art (SOTA) performance.

Particularly, the integration of guidance enables us to apply flow-based models for \emph{return-conditioned plan generating} in offline RL for the first time. The evaluation of offline trained RL agents is via sequential online interactions. The usage of conditional generative models for plan generation~\citep{janner2021offline, janner2022planning, ajay2022conditional} often requires the model to accurately model not only relations within the training data set, but to also generalize to unseen conditional signals during online evaluation. There, we find that guided flows generate reliable execution plans, given the current state and a target return values. Guided Flows also obtain notably higher returns than unguided flows, achieving SOTA performance as well, see Section~\ref{sec:rl_experiments}. 

In addition to its efficacy, for all these aforementioned tasks, Guided Flows also demonstrate favorable compute efficiency and performance tradeoffs. Particularly, for offline RL, Guided Flows enjoys a remarkable 10x speed up compared with diffusion models.


\section{Related Work}
\label{sec:related}

\paragraph{Diffusion and Flow Generative Models.}
Recent major developments in generative models are on building simple models that are highly-efficient to train while providing the means of using conditional inference for solving downstream applications~\citep{janner2022planning,ajay2022conditional, kawar2022denoising, pokle2023training, le2023voicebox}.  
The predominant progress in this domain has primarily centered on diffusion models~\citep{sohl2015deep,ho2020denoising,song2020score}. Within these models, two types of conditional guidance \citep{dhariwal2021diffusion,ho2022classifier} can be employed to promote the conditional information. Whether these form of conditional guidance can be integrated into and help other types of generative models, remains an open question.

As two concurrent works, \citet{dao2023flow} derive an approach similar to ours but is mainly motivated heuristically for the conditional optimal transport probability path; \citet{hu2023latent} develop another different approach by adding an offset to the learned vector field, without theoretical guarantees on the sample distribution. It is noteworthy that we also show the efficacy of guidance in flows across a much wider variety of domain settings.  In particular, our offline RL use case is quite different from the generative modeling tasks considered in those works.




\paragraph{Conditional Generative Modeling in Offline Reinforcement Learning.}
Reinforcement learning (RL) is a powerful paradigm that has been widely applied to solve complex sequential decision making tasks, such as playing games~\cite{silver2016mastering}, controlling robotics~\cite{kober2013reinforcement}, dialogue systems~\cite{li2016deep, singh1999reinforcement}.
The premise of \emph{offline RL} is to learn effective policies solely from static datasets that consist of previous collected experiences, generated by certain unknown policies~\cite{levine2020offline}. There, the agent is not allowed to 
interact with the environment, thus subsides the potential risk and cost of online interactions for high-risk domains such as healthcare.

Conditional generative models are becoming handy tools to help decision making.
A rich body of work~\cite{chen2021decision, janner2021offline, janner2022planning, ajay2022conditional, zhao2023decision, zheng2023semi} focuses on modeling trajectories as sequences of state, action, and reward tokens.  Instead of optimizing the expected return as the classic RL methods, the training objective of all these methods is to simply maximize the likelihood of sequences in the offline dataset. In essence, these methods cast RL as supervised sequence modeling problems, and generative models such as Transformer~\cite{vaswani2017attention,radford2018improving}, Diffusion Models~\cite{ho2020denoising} thus come into play.
Solving RL from this new perspective improves its training stability, and further opens the door of multimodal and multitask pretraining~\cite{zheng2022online, lee2022multi, reed2022generalist}, similar to other domains like language and vision~\cite{radford2018improving,chen2020generative,brown2020language,lu2021pretrained}. There, a generative model can be used as a policy to autoregressively generate actions~\cite{chen2021decision, zheng2022online}; alternatively, it
can generate imagined trajectories which serves as execution plans given current state and a target output such as return or goal location~\cite{janner2021offline, janner2022planning, ajay2022conditional, zhao2023decision}.
There are other works that replace the parameterized Gaussian policies in classic RL methods by generative models to facilitate the modeling of multimodal action distributions~\cite{wang2022diffusion, chi2023diffusion, ward2019improving}. \citet{akimov2022let} apply flow-based models to offline RL as conservative action encoders (and decoders), which decode actions from latent variables output by the policy.

\section{Guided Flow Matching}
\label{sec:guidance}

\paragraph{Setup.} Let $(x_1,y) \sim q(x_1,y)$ denote a true data point where $y\in \Real^k$ is a conditioning variable; $x_0\sim p(x_0)$ is a noise sample, where both data and noise reside in the same euclidean space, \ie, $x_0,x_1\in \Real^d$. Continuous Normalizing Flows (CNFs; \citealt{chen2018neural}) define a map taking the noise sample $x_0$ to a data sample $x_1$ by first learning a vector field $u:[0,1]\times \Real^d\times\Real^k\too \Real^d$ and second, integrating the vector field, \ie, solving the following Ordinary Differential Equation (ODE) 
\begin{equation}\label{e:ode}
    \dot{x}_t = u_t(x_t|y)
\end{equation}
starting at $x_0$ for $t=0$ and solving until time $t=1$. 
For notation simplicity, we use subscript $t$ to denote the first input of function $u$.

\paragraph{Flow Matching (FM).} \citet{lipman2022flow,albergo2022building} propose a method for efficient training of $u$ based on regressing a \emph{target velocity field}. This target velocity field, denoted by $u:[0,1]\times\Real^d\times \Real^k\too\Real^d$, 
takes noise samples $x_0$ and condition vectors $y$ to data samples $x_1$, \ie, it is constructed to generate the following \emph{marginal probability path}
\begin{equation}\label{e:p_t}
    p_t(x|y) = \int p_t(x|x_1) q(x_1|y)dx_1, 
\end{equation}
where $p_t(\cdot|x_1)$ is a probability path interpolating between noise and a single data point $x_1$. That is, $p_t(\cdot|x_1)$ satisfies 
\begin{equation}\label{e:boundary_conds}
 p_0(\cdot|x_1) \equiv p(\cdot), \qquad  p_1(\cdot|x_1) \approx \delta_{x_1}(\cdot),  
\end{equation}
where $\delta_{x_1}(\cdot)$ is the delta probability that concentrates all its mass at $x_1$. Note that
$p_0 \equiv p$ is the noise distribution. An immediate consequence of the boundary conditions in \eqref{e:boundary_conds} is that $p_t(\cdot|y)$ indeed interpolates between noise and data, \ie, $p_0(\cdot|y)\equiv p(\cdot)$ and $p_1(\cdot|y)\approx q(\cdot|y)$ for all $y$.

Next, we assume $u_t(\cdot|x_1)$ is a velocity field that \emph{generates} $p_t(\cdot|x_1)$ in the sense that solutions to \eqref{e:ode}, with $u_t(\cdot|x_1)$ as the velocity field and $x_0\sim p(x_0)$, satisfy $x_t\sim p_t(x_t|x_1)$. The target velocity field for FM is then defined via 
\begin{equation}\label{e:u_t}
    u_t(x|y) = \int u_t(x|x_1) \frac{p_t(x|x_1)q(x_1|y)}{p_t(x)} dx_1,
\end{equation}
and can be proved to generate, in the sense described above, the \emph{marginal probability path} $p_t(\cdot | y)$ in \eqref{e:p_t}. FM is trained by minimizing a tractable loss called the Conditional Flow Matching (CFM) loss (defined later), whose global minimizer  is the target velocity field $u_t$. 

\paragraph{Gaussian Paths.} A popular instantiation of paths $p_t(x|x_1)$ are \emph{Gaussian paths} defined by
\begin{equation}\label{e:gaussian_cond_paths}
 p_t(x|x_1) = \gN(x|\alpha_t x_1, \sigma_t^2 I), 
\end{equation}
where $\gN$ is the Gaussian kernel, $\alpha,\sigma:[0,1]\too [0,1]$ are differentiable functions satisfying $\alpha_0=0=\sigma_1$, $\alpha_1=1=\sigma_0$\footnote{As before, we use subscript $t$ to denote the input of $\alpha$ and $\sigma$.}.  A pair $(\alpha_t,\sigma_t)$ is called a \emph{scheduler}. Marginal paths $p_t(x|y)$  defined with $p_t(x|x_1)$ as in \eqref{e:gaussian_cond_paths} are called \emph{marginal Gaussian paths}.



By convention we denote by $\varnothing$ the null conditioning and set $q(x):=q(x|\varnothing)$,  $\ut_t(x):=\ut_t(x|\varnothing)$, and $u_t(x):=u_t(x|\varnothing)$.
\paragraph{Guided Flows.} Next, we adapt the notion of Classifier-Free Guidance (CFG; \citealt{ho2022classifier}) to conditional velocity fields $u_t(x|y)$. As in CFG, we set our goal to sample from the distribution $\qtilde(x|y) \propto q(x)^{1-\omega} q(x|y)^\omega$, $\omega\in \Real$, 
where only explicit Flow Matching models for $q(x)$ and $q(x|y)$ are given.
Motivated by CFG we define
\begin{equation}\label{e:tilde_ut}
   \tilde{u}_t(x|y) = (1 - \omega) u_t(x) + \omega u_t(x|y)
\end{equation}
To justify this formula for velocity fields, we first relate $u_t(x|y)$ to the
score function $\nabla\log p_t(x|y)$ using the following lemma proved in Appendix \ref{a:lemma_vf_score}: 
\begin{restatable}{lemma}{vfscore}\label{lem:vf_score}
    Let $p_t(x|y)$ be a Gaussian Path defined by a scheduler  $(\alpha_t, \sigma_t)$, then its generating velocity field $u_t(x|y)$ is related to the score function $\nabla\log p_t(x|y)$ by
    \begin{equation}\label{e:lem_u_t}
        u_t(x|y) = a_t x + b_t \nabla\log p_t(x|y),
    \end{equation}
    where 
    \begin{equation}
        a_t=\frac{\dot{\alpha}_t}{\alpha_t}, \qquad b_t= (\dot{\alpha}_t\sigma_t-\alpha_t \dot{\sigma}_t)\frac{\sigma_t}{\alpha_t}.
    \end{equation}    
\end{restatable}
Next, using this the lemma and plugging \eqref{e:lem_u_t} for $u_t(x)=u_t(x|\varnothing)$ and $u_t(x|y)$ in the r.h.s.~of \eqref{e:tilde_ut},
we get
\begin{align}\label{e:tilde_ut_score}
    \tilde{u}_t(x|y) = a_t x + b_t \nabla \log \ptilde_t(x|y),
\end{align}
where
\begin{equation}\label{e:ptilde_t}
   \ptilde_t(x|y) \propto p_t(x)^{1-\omega} p_t(x|y)^{\omega}
\end{equation}
is the geometric weighted average of $p_t(x)$ and $p_t(x|y)$. 

As we prove in Appendix \ref{a:prob_flow_ode}, this velocity field $\tilde{u}_t$ coincides with the one in the Probability Flow ODE~\citep{song2020score} used in Classifier Free Guidance for approximate sampling from distribution $\qtilde(\cdot|y)$. This provides a justification for \eqref{e:tilde_ut}. We note, however, that this analysis shows that both Guided Flows and CFG are guaranteed to sample from $\qtilde(\cdot|y)$ at time $t=1$ if the probability path $\tilde{p}_t(\cdot|y)$ defined in \eqref{e:ptilde_t} is close to the marginal probability path $\int p_t(\cdot|x_1) \qtilde(x_1|y) dx_1$, but 
it is not clear to what extent this assumption holds in practice.



\begin{algorithm}[t]
\DontPrintSemicolon
\small
\caption{Training Guided Flows}
\label{algo:training}
\textbf{Input:} $p_\text{uncond}$ probability of unconditional training\;
Initialize $u_t^\theta$ \; 
\While{not converged}{
$(x_1,y)\sim q(x_1,y)$ {\color{Green}\algorithmiccomment{sample data and condition}}\;
$y\leftarrow \varnothing$ with probability $p_\text{uncond}$ {\color{Green}\algorithmiccomment{null condition}} \;
$x_0\sim p(x_0)$ {\color{Green}\algorithmiccomment{sample noise}}\;
$x_t\leftarrow \alpha_t x_1 + \sigma_t x_0$ {\color{Green}\algorithmiccomment{noisy data point}}\;
$\dot{x}_t\leftarrow \dot{\alpha}_t x_1 + \dot{\sigma}_t x_0$ {\color{Green}\algorithmiccomment{derivative of noisy data point}}\;
Take gradient step on $\nabla_\theta \norm{u^\theta_t(x_t,y) - \dot{x}_t  }^2$
}
\textbf{Output:} $u_t^\theta$
\end{algorithm}

\paragraph{Training Guided Flows.} Training guided flow follows the practice in CFG but replaces the Diffusion training loss with the Conditional Flow Matching (CFM) loss~\citep{lipman2022flow}. This leads to the following loss function:
\begin{equation}\label{e:cfm_loss}
    \resizebox{0.87\columnwidth}{!}{
    $\gL(\theta) = \E_{t, b, q(x_1, y),p(x_0) } \norm{ \ut_t(x_t|(1-b)\cdot y + b \cdot \varnothing) - \dot{x}_t }^2$,
    }
\end{equation}
where $t$ is sampled uniformly in $[0,1]$, 
$b \sim \text{Bernoulli}(p_\text{uncond})$ is used to indicate whether we will use null condition,
$x_1$ and $y$ are sampled from the true data distribution,
$x_t = \alpha_t x_1 + \sigma_t x_0$, $\dot{x}_t=u_t(x_t|x_1)=\dot{\alpha}_t x_1 + \dot{\sigma}_t x_0$ and $\ut:[0,1]\times\Real^d\times\Real^k\too\Real^d$ is a neural network with learnable parameters $\theta\in \Real^p$. The training process is summarized in Algorithm \ref{algo:training}.

\paragraph{Sampling Guided Flows.} Sampling from Guided Flows required approximating the solution to the sampling ODE (\eqref{e:ode}) with the guided velocty field $\tilde{u}_t$ defined in \eqref{e:tilde_ut}, see Algorithm~\ref{algo:sampling_general}. 

\begin{figure}
    \centering
    \begin{subfigure}[b]{0.25\linewidth}
        \includegraphics[width=\linewidth]{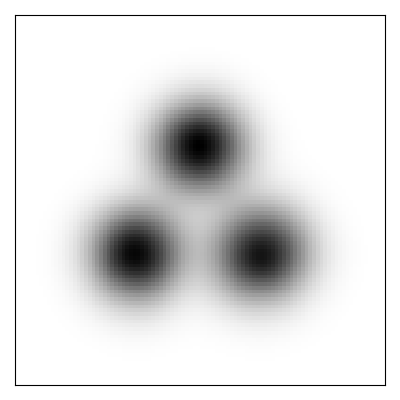}
        \caption*{$w=1.0$}
    \end{subfigure}%
    \begin{subfigure}[b]{0.25\linewidth}
        \includegraphics[width=\linewidth]{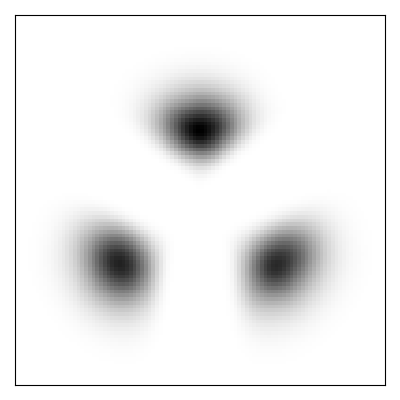}
        \caption*{$w=2.0$}
    \end{subfigure}%
    \begin{subfigure}[b]{0.25\linewidth}
        \includegraphics[width=\linewidth]{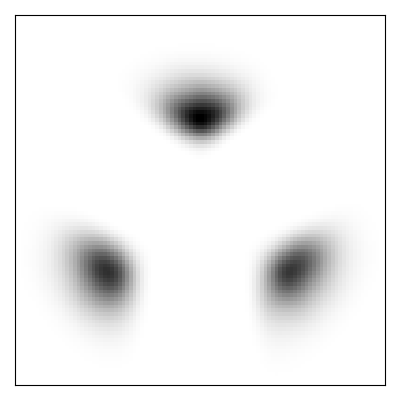}
        \caption*{$w=3.0$}
    \end{subfigure}%
    \begin{subfigure}[b]{0.25\linewidth}
        \includegraphics[width=\linewidth]{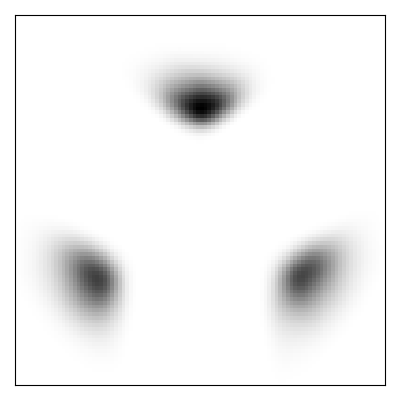}
        \caption*{$w=4.0$}
    \end{subfigure}
    \caption{The effect of increasing guidance weight, conditioned on cluster index. Conditional guidance for all three clusters are shown simultaneously on the same plots.}
    \label{fig:guidance_weight_2d}
\end{figure}

\paragraph{Illustrative Example.} Figure~\ref{fig:guidance_weight_2d} shows a visualization of the effect of Guided Flows in a toy 2D example of a mixture of Gaussian distributions, where $y$ is the latent variable specifying the identity of the mixture component, and  $q(x_1 | y)$ is one single Gaussian component. When the guidance weight is $1.0$, there is no guidance performed, and we are sampling from the unconditional marginal $q(x_1)$.
We see that as guidance weight increases, the samples move away from the unconditional distribution $q(x_1)$.

\begin{algorithm}[t]
\DontPrintSemicolon
\small
\caption{Sampling from Guided Flows}
\label{algo:sampling_general}
\textbf{Input:} trained
velocity field $\ut_t$, condition $y$,
guidance parameter $\omega$, number of ODE steps $\node$\; 
$ x_0 \sim p(x_0)${\color{Green}\algorithmiccomment{sample noise}} \;
$h\leftarrow \frac{1}{\node}$ {\color{Green}\algorithmiccomment{step size}}\;
$\tilde{u}_t(\cdot) \leftarrow (1 - \omega) \ut_t(\cdot) + \omega \ut_t(\cdot|y)${\color{Green}\algorithmiccomment{guided velocity}}\;
\For{$t = 0, h, \ldots, 1-h$}{    
    $x_{t+h} \leftarrow $ \text{ODEStep}($\tilde{u}_t$, $x_t$) {\color{Green}\algorithmiccomment{ODE solver step}}\;
}
\textbf{Output:} $x_1$
\end{algorithm}

\section{Conditional Generative Modeling}
\label{sec:image}
In this section, we perform experiments to test whether Guided Flows can help achieve better sample quality than unguided models. In particular, we consider two application settings: conditional image generation and zero-shot text-to-speech synthesis. The goal of these experiments to test Guided Flows on standard generative modeling settings, where the evaluation is purely based on sample quality, while exploring different data modalities, before we move on validating Guided Flows for more complex planning tasks in \Cref{sec:rl}.

\subsection{Conditional Image Generation}

\begin{figure*}
    \centering
    
    \begin{subfigure}[b]{0.02\linewidth}
        \raisebox{1.1em}{\rotatebox[origin=t]{90}{1.0}}
    \end{subfigure}%
    \begin{subfigure}[b]{0.48\linewidth}
        \caption*{Brambling}
        \includegraphics[width=\linewidth]{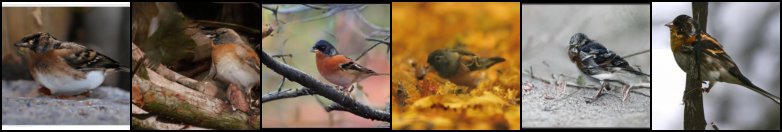}
    \end{subfigure}
    \begin{subfigure}[b]{0.48\linewidth}
        \caption*{Husky}
        \includegraphics[width=\linewidth]{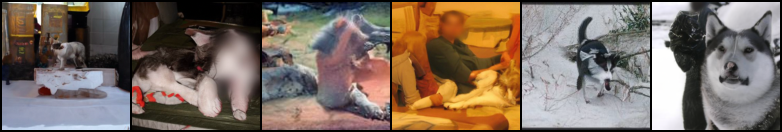}
    \end{subfigure}\\
    \begin{subfigure}[b]{0.02\linewidth}
        \raisebox{1.1em}{\rotatebox[origin=t]{90}{2.0}}
    \end{subfigure}%
    \begin{subfigure}[b]{0.48\linewidth}
        \includegraphics[width=\linewidth]{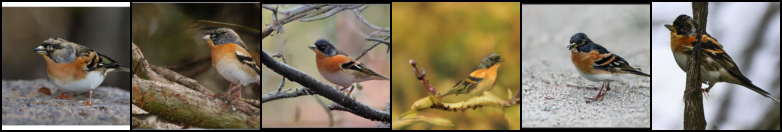}
    \end{subfigure}
    \begin{subfigure}[b]{0.48\linewidth}
        \includegraphics[width=\linewidth]{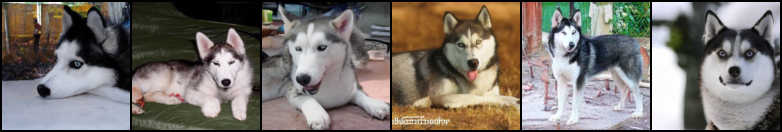}
    \end{subfigure}\\
    \begin{subfigure}[b]{0.02\linewidth}
        \raisebox{1.1em}{\rotatebox[origin=t]{90}{3.0}}
    \end{subfigure}%
    \begin{subfigure}[b]{0.48\linewidth}
        \includegraphics[width=\linewidth]{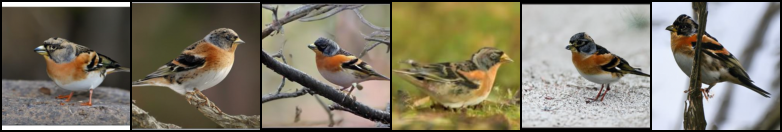}
    \end{subfigure}
    \begin{subfigure}[b]{0.48\linewidth}
        \includegraphics[width=\linewidth]{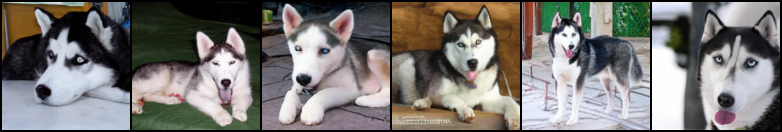}
    \end{subfigure}

    \vspace{0.5em}

    \begin{subfigure}[b]{0.02\linewidth}
        \raisebox{1.1em}{\rotatebox[origin=t]{90}{1.0}}
    \end{subfigure}%
    \begin{subfigure}[b]{0.48\linewidth}
        \caption*{Meerkat}
        \includegraphics[width=\linewidth]{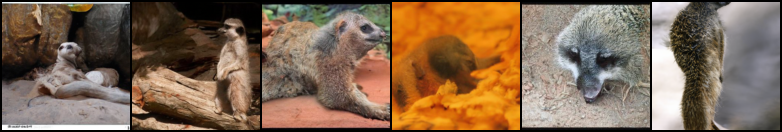}
    \end{subfigure}
    \begin{subfigure}[b]{0.48\linewidth}
        \caption*{Otter}
        \includegraphics[width=\linewidth]{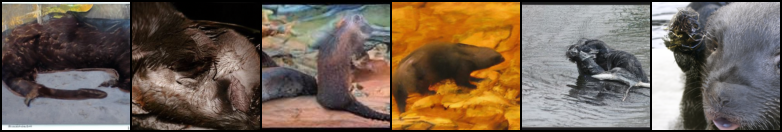}
    \end{subfigure}\\
    \begin{subfigure}[b]{0.02\linewidth}
        \raisebox{1.1em}{\rotatebox[origin=t]{90}{2.0}}
    \end{subfigure}%
    \begin{subfigure}[b]{0.48\linewidth}
        \includegraphics[width=\linewidth]{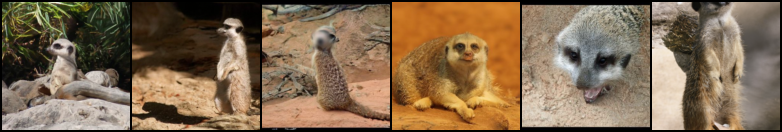}
    \end{subfigure}
    \begin{subfigure}[b]{0.48\linewidth}
        \includegraphics[width=\linewidth]{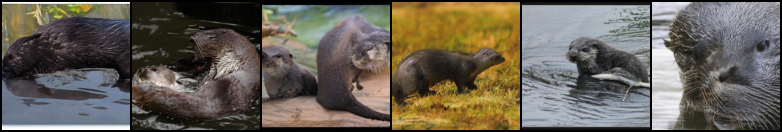}
    \end{subfigure}\\
    \begin{subfigure}[b]{0.02\linewidth}
        \raisebox{1.1em}{\rotatebox[origin=t]{90}{3.0}}
    \end{subfigure}%
    \begin{subfigure}[b]{0.48\linewidth}
        \includegraphics[width=\linewidth]{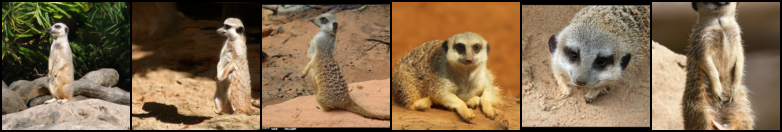}
    \end{subfigure}
    \begin{subfigure}[b]{0.48\linewidth}
        \includegraphics[width=\linewidth]{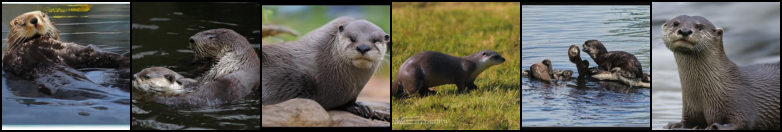}
    \end{subfigure}
    
    \caption{Generated samples from face-blurred ImageNet using four different class labels, showing the effect of different guidance weights with Guided Flows. Higher guidance weights result in more class-specific features to appear.}
    \label{fig:imagenet_samples}
\end{figure*}

We downsample the official \emph{face-blurred} ImageNet dataset to images of 64×64 pixels, using the open
source preprocessing scripts from \citet{chrabaszcz2017downsampled}.
We train Guided Flow Matching models $p_t(x | y)$ where $x$ denotes an image and $y$ is the class label of that image. In particular, we consider two affine Gaussian probability paths, the optimal transport (FM-OT) path considered by \citet{lipman2022flow} and the cosine scheduling (FM-CS) path considered by \citet{albergo2022building}. 
As a baseline, we train diffusion models (DDPM; \citealt{ho2020denoising, song2020score}) with classifier-free guidance \citep{ho2022classifier}. We report the results of both standard sampling of DDPM and also the deterministic DDIM sampling algorithm \citep{song2020denoising}. All the models have the same U-Net architecture adopted from \citet{dhariwal2021diffusion}, and trained with the same hyperparameters and number of iterations, as listed in Table~\ref{tbl:imagenet_hp}.

\begin{figure}[t]
    \centering
    \begin{subfigure}[b]{0.5\linewidth}
    \includegraphics[width=1.0\linewidth]{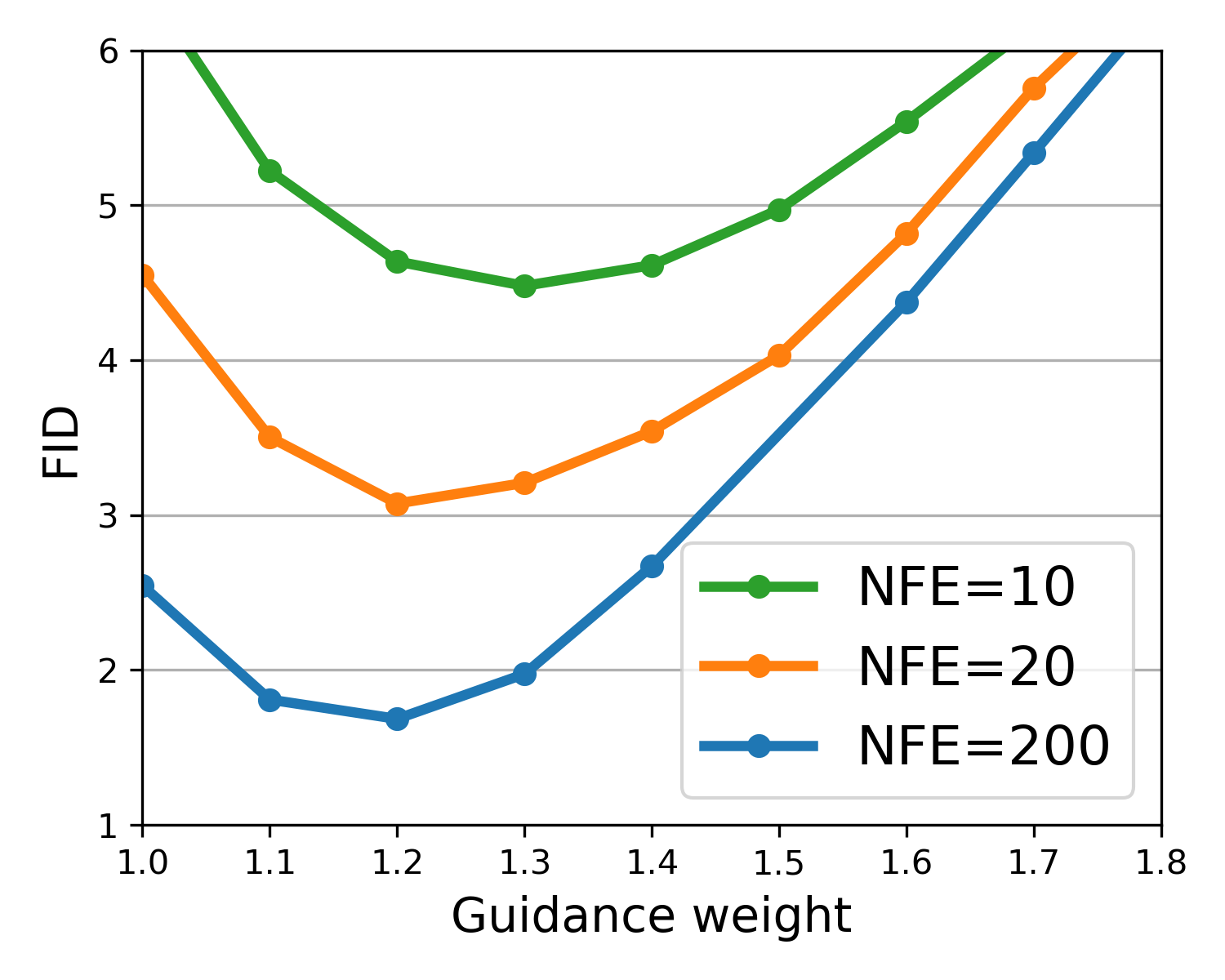}
    \caption{Guidance}
    \end{subfigure}%
    \begin{subfigure}[b]{0.5\linewidth}
    \includegraphics[width=1.0\linewidth]{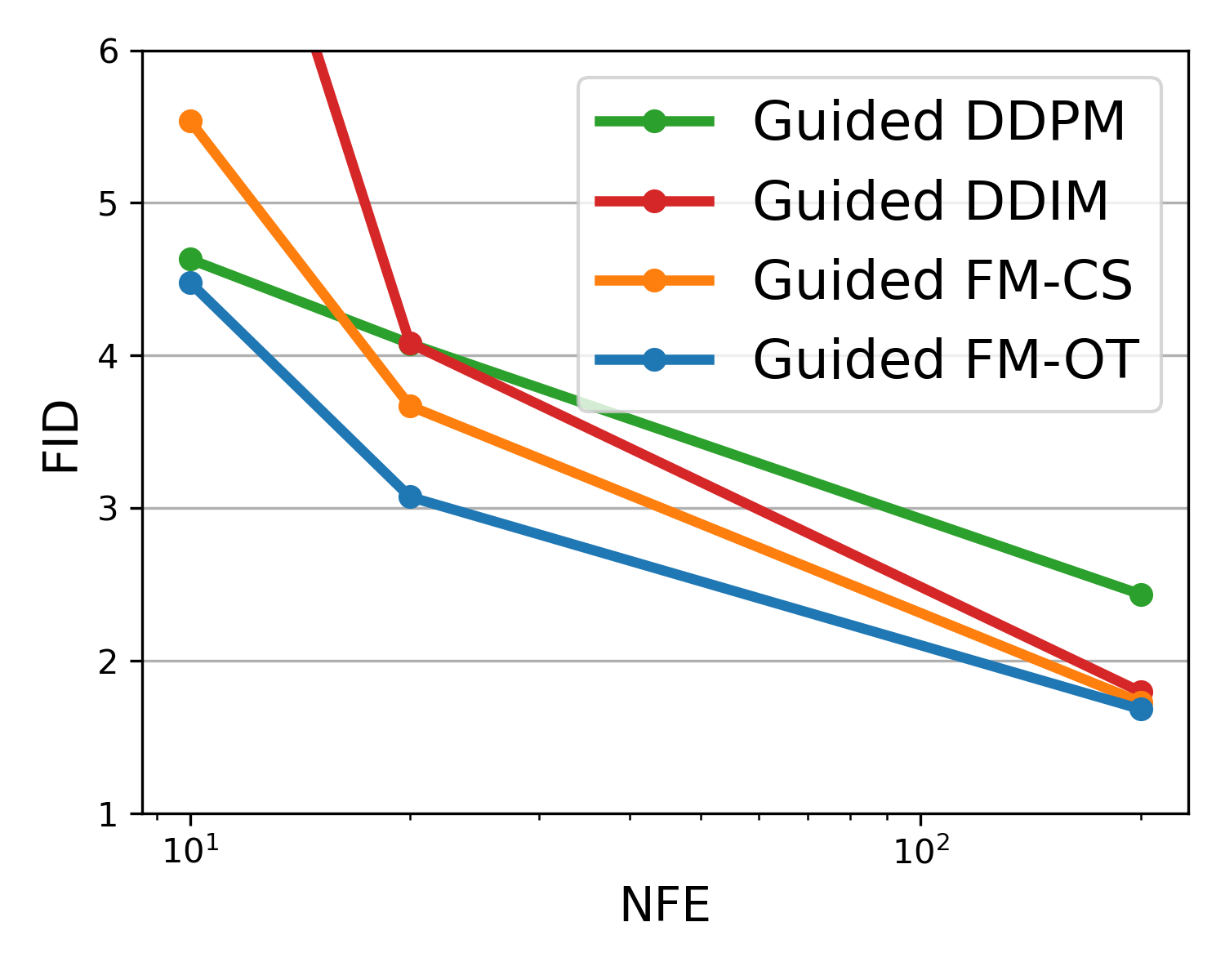}
    \caption{Efficiency}
    \label{fig:imagenet64_efficiency}
    \end{subfigure}
    \caption{(a) The performance of Guided Flows (FM-OT) as the guidance weight $\omega $ changes. When $\omega = 0$, 
    there is no guidance performed; with $\omega > 1.0$, Guided Flows (FM-OT) achieve a lower FID. This trend is consistent when we use different NFEs (ODE steps). With NFE=200, the guided flow achieves FID $1.68$ on ImageNet-64.(b) Guided Flows with conditional optimal transport paths have the best tradeoff for efficiency, compared with other baselines.} 
    \label{fig:imagenet64} 
\end{figure}

Results are displayed in \Cref{fig:imagenet64}. We see that guidance for Flow Matching models can drastically help increase sample quality (reducing FID from $2.54$ to $1.68$) using a midpoint solver with 200 number of function evaluations (NFE), i.e. 200 ODE steps. We do see that the optimal guidance weight can change depending on the compute cost (NFE), with NFE=10 having a higher optimal guidance weight. Additionally, in \Cref{fig:imagenet64_efficiency} we show the Pareto front of the sample quality and efficiency tradeoff, plotted for each model.
We note that the optimal guidance weight can be very different between each model, with DDPM noticeably requiring a larger guidance weight.
Here we find that FM-OT models are slightly more efficient than the other models that we consider.

\subsection{Zero-shot Text-to-Speech Synthesis}

\begin{table}[t]
    \centering
    \setlength{\tabcolsep}{4.5pt}
    \begin{tabular}{ccc}
    \toprule
    Guidance Weight & Continuation & Text-only \\
    \midrule
    \citet{le2023voicebox} & 2.0\hphantom{0} & 3.1\hphantom{0} \\
    \hdashline
    \addlinespace[2pt]
    \textcolor{gray}{1.0} & \textcolor{gray}{2.06} & \textcolor{gray}{3.09} \\ 
    1.2 & 1.95 & 2.98 \\
    1.4 & 1.96 & 2.89 \\
    1.6 & 1.95 & 2.83 \\
    1.8 & \highlight{1.94} & 2.87 \\
    2.0 & 1.98 & 2.82 \\
    2.2 & 1.96 & 2.80 \\
    2.4 & 2.10 & 2.83 \\
    2.6 & 2.40 & 2.76 \\
    2.8 & 3.40 & 2.79 \\
    3.0 & 5.10 & \highlight{2.75} \\
    \bottomrule
    \end{tabular}
    \caption{Word error rates (WER) for zero-shot text-to-speech. When the weight is $1.0$, there is no guidance performed and we sample directly from the trained conditional distribution. We see that performance improves when using Guided Flows with weights greater than 1.0.}
    \label{tab:speech}
\end{table}

Given a target text and a transcribed reference audio as conditioning information $y$, zero-shot text-to-speech (TTS) aims to synthesize speech resembling the audio style of the reference, which was never seen during training. As our Guided Flow model, we train a model on 60K hours ASR-transcribed English audiobooks, following the experiment setup in \citet{le2023voicebox}.

Specifically, we consider two main tasks. The first one is zero-shot TTS where the first 3 seconds of each utterance is provided and the model is requested to continue the speech. The second one is diverse speech generation, where only the text is provided to the model. In order to assess the accuracy of the generated results, we report the word error rate (WER) using automatic speech recognition (ASR) models following prior works~\citep{wang2018style}.

Results of Guided Flows is provided in \Cref{tab:speech}, where we also provide the results of \citet{le2023voicebox} as reference. We also report our results when no guidance is used (weight equal to 1.0). With guidance, we see a marginal improvement for the continuation TTS task, and a much more sizable gain in the text-only TTS task. This likely due to the text-only TTS task being a much more diverse distribution.
\section{Planning for Offline RL}
\label{sec:rl}
\subsection{Preliminaries}
We model our environment as a Markov decision process (MDP)~\citep{bellman1957mdp} denoted by $\langle \S, \A, p, P, R, \gamma \rangle$, where $\S$ is the state space, $\A$ is the action space, $p(s_0)$ is the distribution of the initial state, $P(s_{t+1}|s_t, a_t)$ is the transition probability distribution, $R(s_t, a_t)$ is the deterministic reward function, and $\gamma$ is the discount factor. At timestep $t$, the agent observes a state $s_t \in \S$ and executes an action $a_t \in \A$.
The environment will provide the agent with a reward $r_t = R(s_{t}, a_t)$, and also moves it to the next state $s_{t+1} \sim P(\cdot| s_t, a_t)$.
Let $\tau$ be a trajectory. For any length-$H$ subsequence $\tausub$ of $\tau$, e.g., from timestep $t$ to $t+H-1$, we define the return-to-go (RTG) of $\tausub$ to be the sum of its discounted return $g(\tausub) = \sum_{t' = t}^{t+H-1} \gamma^{t'-t} r_{t'}$\footnote{This is slightly different from the standard RTG definition where the discounting factor is $\gamma^{t'-t}$ rather than $\gamma^{t}$.}.  We also use $\vs(\tausub)$ to denote the state sequence extracted from the subsequence $\tausub$.
A deterministic inverse dynamics model (IDM) is a function $f: \S \times \S \mapsto \A$ which predicts action using states: $\ahat_t = f(s_t, s_{t+1})$.

\subsection{Our Setup}
\begin{savenotes}
\begin{algorithm}[t]
\DontPrintSemicolon
\small
\caption{A General Framework of Conditioned Generative Planning}
\label{algo:planning}
\textbf{Input:} trained conditional sequence model $p_{\theta}(s_{t+1}, \ldots, s_{t+H-1} | s_t, g_t)$,
IDM $f_\rho(s, s')$, 
initial state $s_0$,
initial conditioning parameter $g_0$, 
conditioning parameter updating rule $G$\;
$t \leftarrow 0$\;
\While{episode not done}{
   Sample $(\shat_{t+1}, \ldots, \shat_{t+H-1}) \sim p_\theta(\cdot | s_t, g_t)$\footnote{As reflected in Figure~\ref{fig:return_conditioned_seq_model_train} and~\ref{fig:return_conditioned_seq_model_eval}, we model the sequence starting from
$s_t$ rather than $s_{t+1}$ in the actual implementation. Due to the presence of $s_t$, the sampling process is slightly adjusted from Algorithm~\ref{algo:sampling_general}, as we need to zero out the vector fields corresponding to $s_t$. See Algorithm~\ref{algo:sampling_rl}. 
}\; 
   Predict action $\ahat_t = f_\rho(s_t, \shat_{t+1})$\;
   Execute $\ahat_t$ and observe $s_{t+1}$ and $r_t$\;
   $g_{t+1} \leftarrow $ compute the next conditioning parameter according to $G$\;
   $t \leftarrow t+1$\;
}
\end{algorithm}
\end{savenotes}

\begin{figure*}[t]
    \centering
    \includegraphics[width=0.4\textwidth]{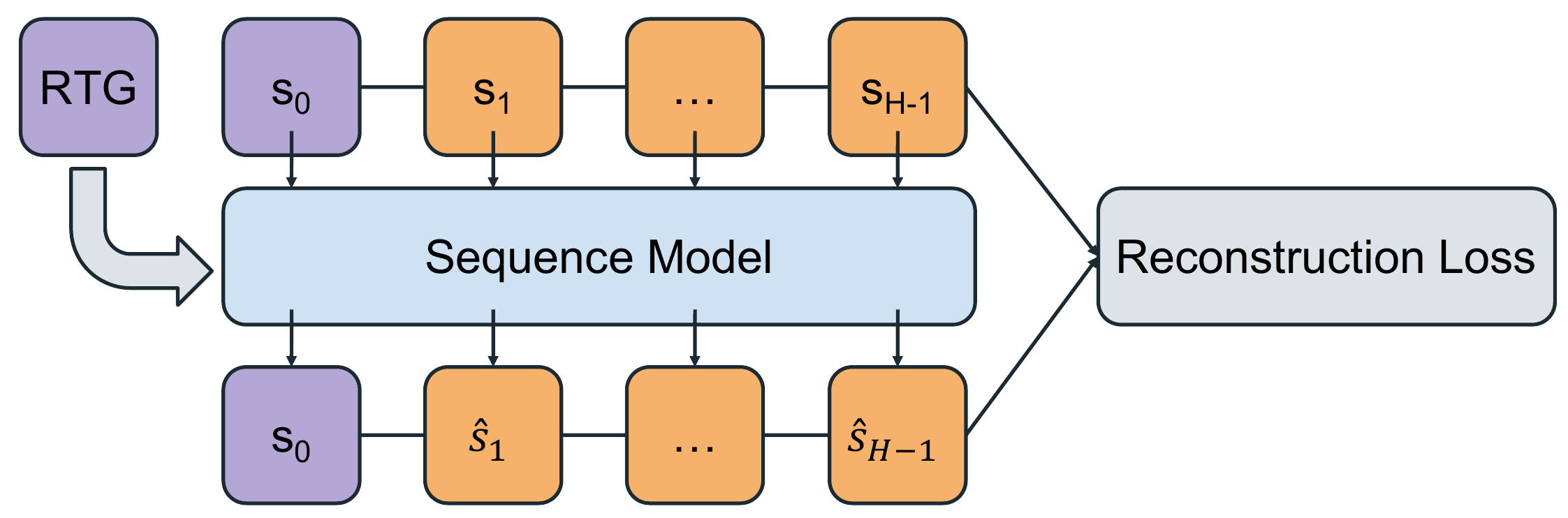}
    \hskip0.1\textwidth
    \includegraphics[width=0.33\textwidth]{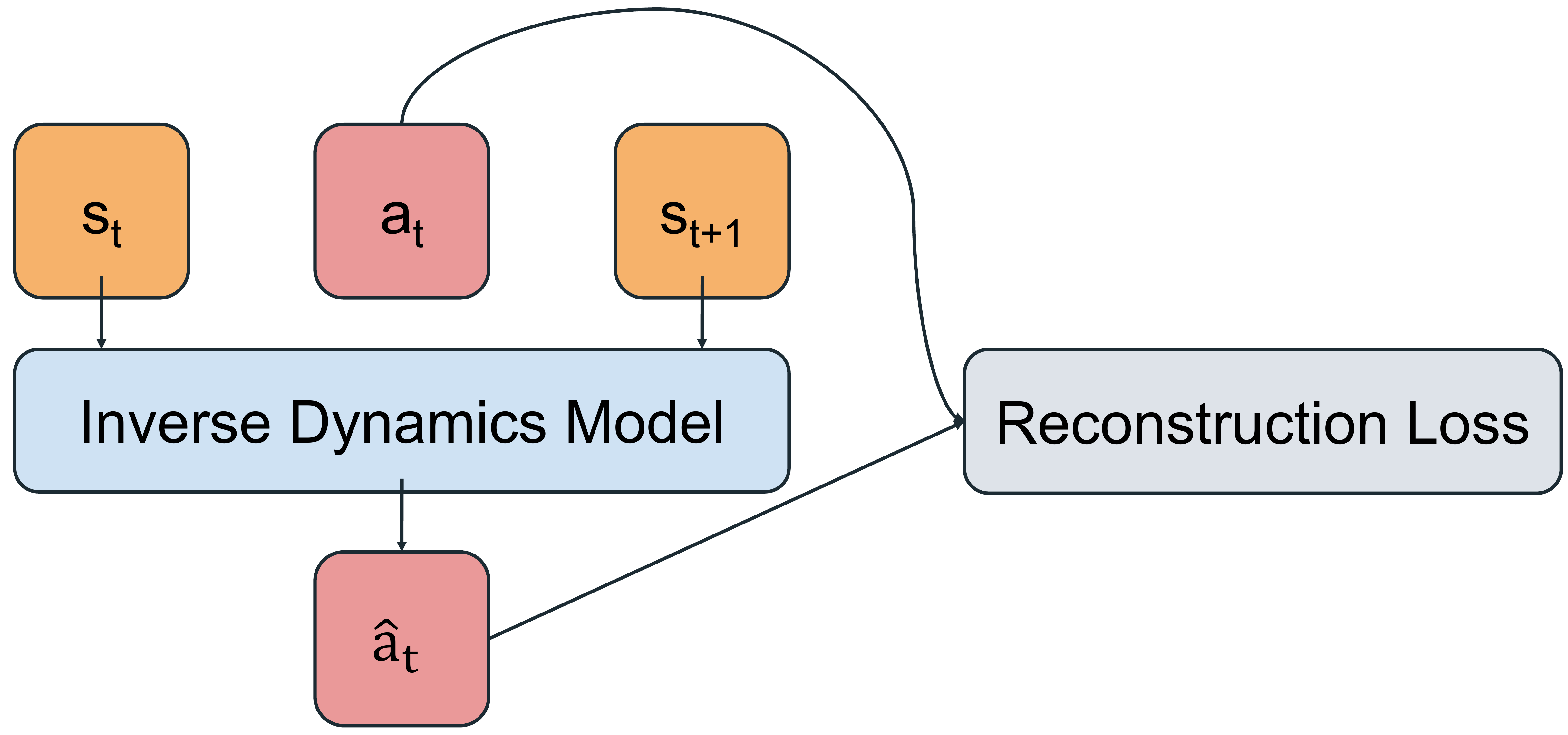}
    \caption{The training phase of return-conditioned generative planning. The state sequence model is trained in a classifer-free manner---with certain probability, it is conditioned on the RTG.}
    \label{fig:return_conditioned_seq_model_train}
\end{figure*}
\begin{figure}[t]
\small
\centering
     \includegraphics[width=0.5\columnwidth]{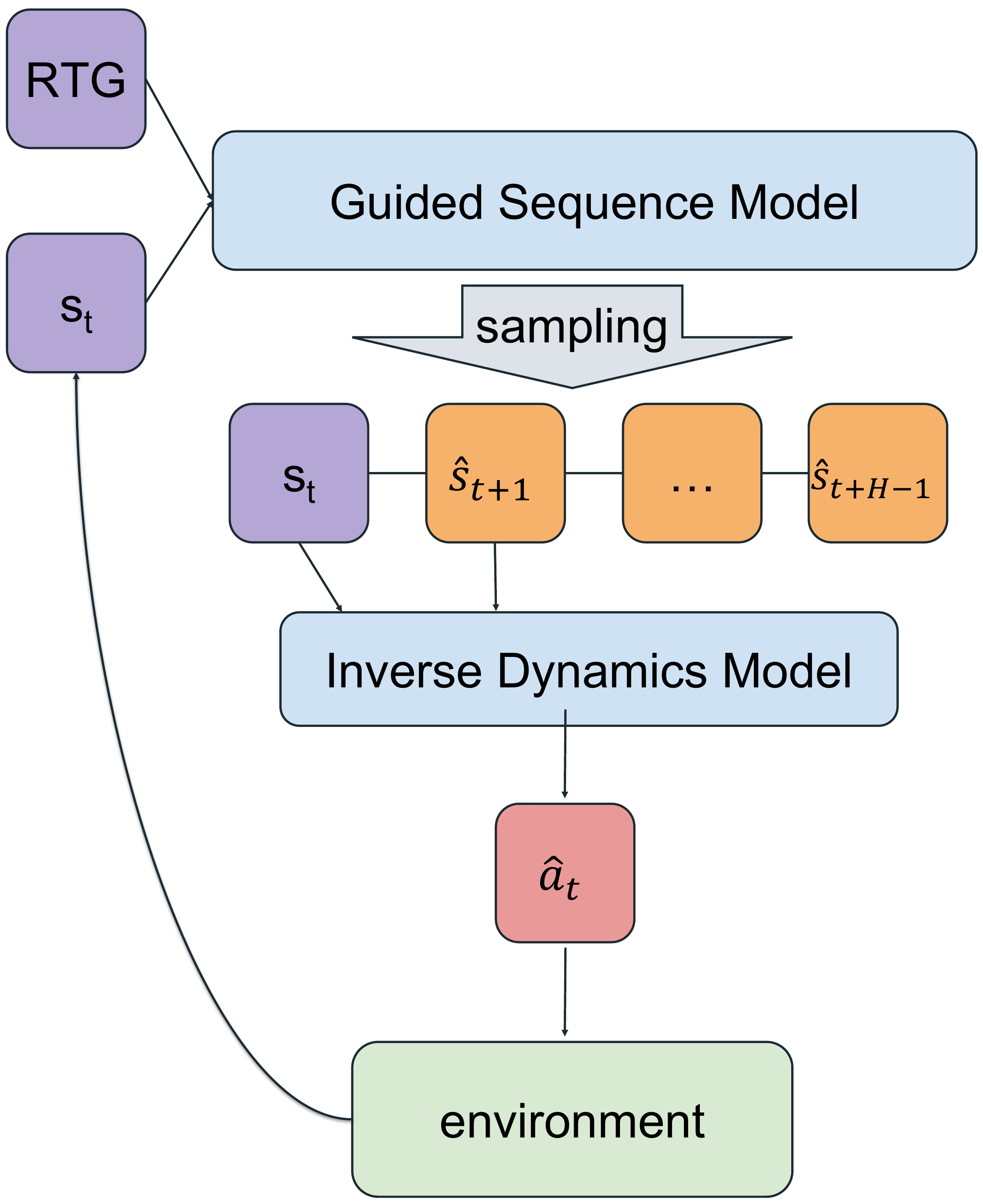}
    \caption{The evaluation phase using return-conditioned generative planning.}
    \label{fig:return_conditioned_seq_model_eval}   
\end{figure}
We consider the paradigm proposed by~\citet{ajay2022conditional}, where a generative model learns to predict a sequence of future states, conditioning on the current state and target output such as expected return. Intuitively, the predicted future states form a \emph{plan} to reach the target output, and the predicted next state can be interpreted as the \emph{next intermediate goal} on the roadmap. Based on the current state and the predicted next state, we use an inverse dynamics model to predict the action to execute. 
Algorithm~\ref{algo:planning} summarizes this framework. 
We emphasize that the \emph{target output $g_t$ is the conditioning variable where guidance will perform},
and the current state $s_t$ is a general conditioning variable.
In this work, we consider the target return as our conditioning variable. Return-conditioned RL methods are widly used to solve 
standard RL problems where the environment produces dense rewards~\cite{srivastava2019training, kumar2019reward, schmidhuber2019reinforcement, emmons2021rvs, chen2021decision, nguyen2022reliable}. 
Figure~\ref{fig:return_conditioned_seq_model_train} and~\ref{fig:return_conditioned_seq_model_eval} plot the training and evaluation phases of our paradigm.
Careful readers might notice that we resample the whole sequence $\hat{s}_{t+1}, \ldots, \hat{s}_{t+H-1}$ at every timestep $t$,
but only use $\hat{s}_{t+1}$ to predict $\hat{a}_t$. In fact, it is completely feasible to predict the actions for the next multiple steps using a single plan, which also saves computation. We note that replanning at every timestep is for the sake of planning accuracy, as the error will accumulate as the horizon expands, and there is a tradeoff between computational efficiency and agent performance. For diffusion models, previous works have used heuristics to improve the execution speed, e.g., reusing previously generated plans to warm-start the sampling of subsequence plans~\cite{janner2022planning, ajay2022conditional}. This is beyond the scope of our paper, and we only consider replanning at every timestep for simplicity.

\subsection{Experiments}
\label{sec:rl_experiments}
Our experiments aim at answering the following questions:
\begin{enumerate}[leftmargin=3em]\itemsep0em
    \item[Q1.] Can conditional flows generate meaningful plans for RL problems given a target return?
    \item[Q2.] How do flows compare to diffusion models, in terms of both downstream RL task performance and compute efficiency?
    \item[Q3.] Compared with unguided flows, can guidance help planning?
\end{enumerate}

\paragraph{Tasks and Datasets}
We consider three \gym locomotions tasks, \hopper, \walker and \cheetah,
using offline datasets from the D4RL benchmark~\cite{fu2020d4rl}.
For all the experiments, we train 5 instances of each method with different seeds.
For each instance, we run 20 evaluation episodes. We shall discuss 
the model architectures and important hyperparameters below,
and we refer the readers to Appendix~\ref{app:expr_details} for more details.

\paragraph{Baseline} We compare our method to Decision Diffuser~\cite{ajay2022conditional},
which uses the same paradigm with diffusion models to model the state sequences.

\paragraph{Sequence Model Architectures}
For the locomotion tasks, we train flows for state sequences of
length $H=64$, parameterized through the velocity field
$\ut$. Similar to the previous work~\cite{janner2022planning, ajay2022conditional},
the velocity field $\ut$ is modeled as a temporal U-Net consisting
of repeated convolutional residual blocks. Both the time $t$ and the RTG $g(\vs)$
are projected to latent spaces via multilayer perceptrons, where $t$ is first transformed to
its sinusoidal position encoding. 
For the baseline method, the diffusion model is training similarly with $200$ diffusion steps, 
where we use a temporal U-Net to predict the noise at each diffusion steps 
throughout the diffusion process. 

\paragraph{Inverse Dynamic Model} For all the environments and datasets, 
we model the IDM by an MLP with 2 hidden layers and 1024 hidden units per layer. 
Among all the offline trajectories, we randomly sample 10\% of them as the validation set.
We train the IDM for 100k iterations,  and use the one that yields the best validation performance.

\paragraph{Low Temperature Sampling} To sample from the diffusion model, we use $200$ diffusion steps. 
For the sake of fair comparison, we also use $200$ ODE steps when sampling from the flow matching model.
Following \citet{ajay2022conditional}, we use the \emph{low temperature sampling} technique for diffusion model, 
where at each diffusion step $k$ we sample the state sequence from $\N(\hat{\mu}_k, \alpha^2 \hat{\Sigma}_k)$\footnote{$\hat{\mu}_k$ and $\hat{\Sigma}_k$ 
are the predicted mean and variance for sampling at the $k$th diffusion step. We refer the readers to \citet{ho2020denoising} for more details.} with a hand-selected temperature parameter $\alpha \in (0, 1)$. 
We sweep over 3 values of $\alpha$ for all our experiments: $0.1, 0.25$, and $0.5$.
For flow matching, we analogously set the initial distribution $p_0(x_0) = \N(0, \nu^2 I)$ and we sweep over two values of $\nu$: $0.1$ and $1$.

\begin{table*}[t]
    \centering
\resizebox{\textwidth}{!}{
\begin{tabular}{c c c c c c c c c c c}
    \toprule \hline 
         & \multicolumn{3}{|c|}{\hopper} & \multicolumn{3}{c|}{\walker} & \multicolumn{3}{c|}{\cheetah} & \multirow{2}{*}{Average}\\ 
         & \medreplay & \med &\medexpert & \medreplay & \med &\medexpert & \medreplay & \med &\medexpert & \\ \midrule
   Flow      & \highlight{0.89} &	\highlight{0.84} &	1.05 & \highlight{0.78} &	0.77&	0.94& 0.42&	\highlight{0.49}&	\highlight{0.97} & \highlight{0.79}\\ 
   Diffusion & 0.87 &	0.72 &	\highlight{1.09} & 0.64	& \highlight{0.8}	&\highlight{1.07} &\highlight{0.48}	&0.41	&0.95 & 0.78\\
   \hline \bottomrule 
    \end{tabular}
}
    \caption{The normalized return obtained when the state sequence models are trained by flows and diffusion models. We use the same IDM for both methods. Results aggregated over 5 training instances with different seeds.}
    \label{tab:expr_rl_locomotion}
\end{table*}

\subsubsection{Plan Generation (Q1)}
\begin{figure}[h]
    \centering
    \includegraphics[width=\columnwidth]{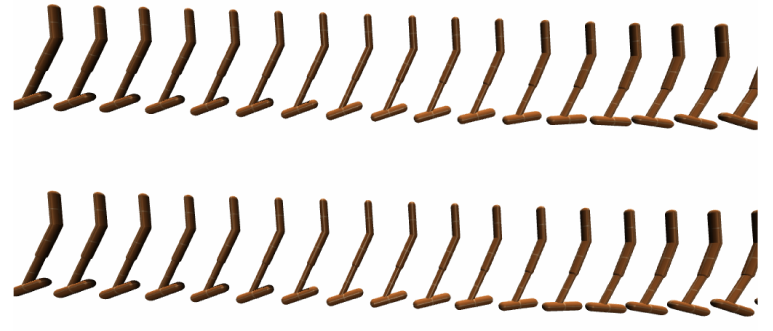} 
    \caption{The top panel plots a truth state sequence $s_0, s_,1 \ldots$ randomly sampled from the \hopper-\medium dataset. The bottom panel plots the flow generated state sequence, conditioning on the first state $s_0$ with guidance weight $3.0$. These two sequences are very similar, demonstrating that guided flow is capable to generate meaningful plans to navigate the agent.}
    \label{fig:hopper_traj}
\end{figure}

To verify the capability of flows to generate meaningful plans that can guide the agent,
we sample from a flow on the \hopper-\medium dataset. We randomly select a subtrajectory $\tausub$
from the dataset, and let flow condition on its first state and RTG $0.8$. 
Figure~\ref{fig:hopper_traj} plots both state sequences. We can see that the generated state sequence is almost identical
to the ground truth, demonstrating that guided flow is capable to generate meaningful plans to navigate the agent.
We note that this RTG value $0.8$ is out-of-distribution (OOD), as the maximum RTG value of the training dataset is $0.61$.
This suggests that guided flows might be even robust to OOD RTG values\footnote{We note that the fundamental task for offline RL is to address the offline-to-online distribution shift.
During online evaluation, an offline trained agent might encounter unseen data, potentially resulting in the generation of 
unreasonable actions or states that lead to poor performance.
To address this issue, various notions of conservatism has been introduced into offline RL algorithms. The overarching objective of those diverse
conservatism techniques is to maintain the output of the algorithm close to the training data distribution.
}, which we believe is an interesting property to understand
and a potential direction to explore for future work, see related discussions in \citet{chen2021decision, emmons2021rvs, zheng2022online, nguyen2022reliable}.
We refer the readers to Figure~\ref{fig:hopper_trajs_more} for more examples of generated plans. 

\begin{figure*}[t]
    \centering
    \includegraphics[width=0.75\textwidth]{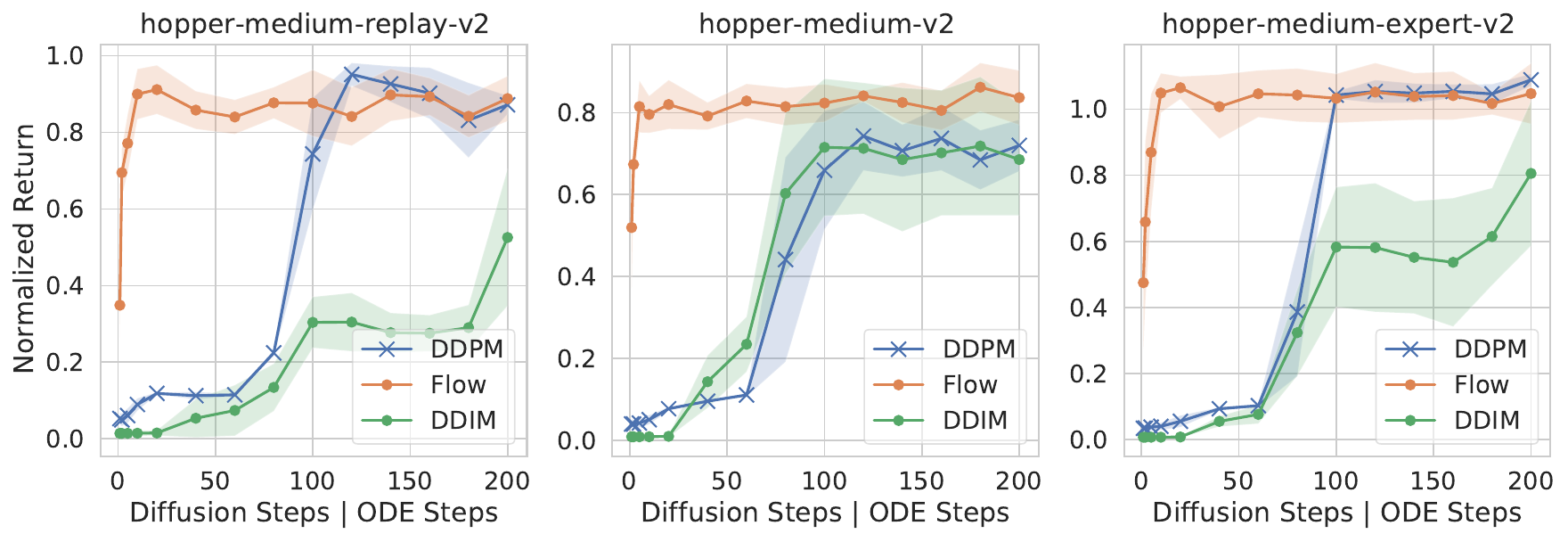} 
    \caption{\small The obtained normalized return when the internal samplling steps used to generate the state sequence varies.}
    \label{fig:locomotion_diffusion_step_comparison}
\end{figure*} 
\subsubsection{Benchmark (Q2)}
In this section, we conduct experiments to investigate the efficacy and efficiency of guided flows. We highlight the predominant trends from our findings:

{
\centering
    \emph{\hskip15pt Guided flows are on par with diffusion models with respect to the absolute performance,
    with a significant 10x speed up in sampling.}
}

\paragraph{Performance Efficacy} Throughout all our experiments, in addition to the temperature parameter for sampling, for both methods, we 
sweep over 3 values of RTG: 0.4, 0.5, and 0.6 for the \medreplay and \med datasets of \cheetah, and 0.7, 0.8 and 0.9 for all the other datasets.
We also sweep over 4 values of guidance parameters: $1.8, 2.0, 2.2, 2.4$ for diffusion models, and $1.0, 1.5, 2.0, 2.5$ for flows.
We train both guided flows and guided diffusion models for 2 million iterations, and save checkpoints every $200k$ iteration.
We evaluate the performance on all the saved checkpoints and report the best result in Table~\ref{tab:expr_rl_locomotion}\footnote{This is because \citet{ajay2022conditional} report the best results over a collection of checkpoints.}. The performances of flow matching and diffusion model
are comparable, where flow matching performs marginally better. \looseness=-1

\paragraph{Computational Efficiency}
A well-known pain-point of diffusion model is its computational inefficiency,
since sampling from a diffusion model consists of iterative denoising steps.
In our case, the diffusion model is trained with $200$ diffusion steps, thus it takes 
$200$ internal sampling steps to generate a sample with full quality. To accelerate sampling,
many algorithms have been proposed to reduce the number of internal steps, including implicit models with deterministic sampling (DDIM, ~\citealt{song2020denoising}),
distillation~\cite{salimans2022progressive},
noise schedule~\cite{chen2020wavegrad, nichol2021improved, lin2023common}.
As a tradeoff, these methods all lead to loss in sample quality. Similarly, sampling from a flow model requires a number of internal steps to solve the ODE~\eqref{e:ode}, and more internal steps leads to sample with better quality.

\begin{figure}[ht]
    \centering
    \includegraphics[width=0.55\columnwidth]{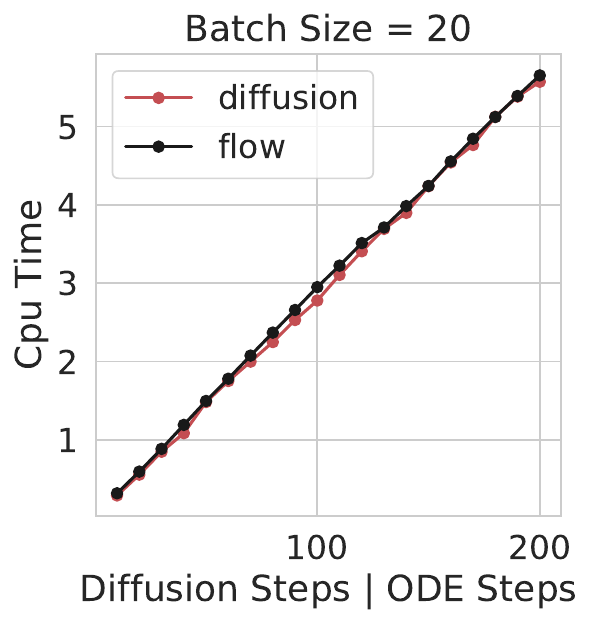}
    \caption{\small The CPU time of generating a batch of $704$-dimensional vectors (the subsequence length is 64, and
each state is of dimension 11), where the batch size is $20$. 
Results averaged over $100$ replications. }
    \label{fig:locomotion_time_comparison}
\end{figure}

To understand the computation-vs-quality tradeoff for both flows and diffusion models, we 
run an ablation experiment for the \hopper task,
where we only sample with $K$ internal steps $(K \leq 200)$ for both methods. In addition to the standard diffusion model (DDPM, \citet{ho2020denoising}), we also compare with DDIM~\cite{song2020denoising}, a deterministic sampling algorithm widely used in diverse domains. Figure~\ref{fig:locomotion_diffusion_step_comparison}
plots the normalized return we obtain versus the number of internal steps. For all 3 datasets, phase transitions occur for all three methods. Surprisingly, $10$ ODE steps are sufficient for flows to generate
samples leading to the same return as $200$ ODE steps; whereas DDPM and DDIM both need $100$ diffusion steps. Again, the performance of flows
is comparable to DDPM and is better than DDIM. Next, we compare the CPU time consumed by these methods. 
Figure~\ref{fig:locomotion_time_comparison} shows that the CPU time
consumed by the diffusion model and the flow are roughly the same when the number of internal steps match, and it scales linearly as the number of internal steps increase. This means, compared with diffusion model, flows only need $10\%$ computing time to 
generate samples leading to the same downstream performance.  The trend remains the same when the batch size increase, see Figure~\ref{fig:locomotion_time_comparison_all_batchsizes}. \looseness=-1



\subsubsection{Influence of Guidance (Q3)}
Table~\ref{tab:guidance_rl} reports the normalized return obtained by our agents
for the \hopper and \cheetah tasks, where the guidance weight varies from 1.0 to 3.0. 
The flows are trained on both \med and \medreplay datasets for \hopper, and
the \med dataset for \cheetah. The guidance weight $1.0$ yields unguided flows.
The results show that the guided flows outperform the unguided ones on all three datasets.
\begin{table}[ht]
    \centering
    \setlength{\tabcolsep}{4.5pt}
    \resizebox{\columnwidth}{!}{
        \begin{tabular}{cccc}
        \toprule
        \multirow{2}{*}{Guidance Weight}  & \multicolumn{2}{c}{\hopper}  & \cheetah\\
        & \medium & \medreplay & \medium \\ 
        \midrule
        \textcolor{gray}{1.0} & \textcolor{gray}{0.64} & \textcolor{gray}{0.83} &  \textcolor{gray}{0.48}\\ 
        1.5 & 0.73 & 0.87 &0.48\\
        2.0 & 0.81 & \highlight{0.89} &\highlight{0.49}\\
        2.5 & 0.81 & 0.86 & 0.46\\
        3.0 & \highlight{0.84}& 0.84  & 0.45\\
        \bottomrule
        \end{tabular}
    }
    \caption{The normalized return obtained when using different guidance weights. For both tasks, guided flows outperform unguided flows (guidance weight is $1$). }
    \label{tab:guidance_rl}
\end{table}

\section{Conclusion}
\label{sec:conclusion}
We thoroughly explore the theory and effect of guidance for flow matching. We empirically validate the conditional generative capabilities of flow-based models trained through recently-proposed simulation-free algorithms~\citep{lipman2022flow,albergo2022building} for a variety of applications, confirming its success across diverse domains. Our experiments show that conditional guidance can lead to better results for flow-based models. Moreover, guided flows excel in standard generative tasks like image synthesis and speech generation, achieving SOTA performance.  Additionally, our experiments highlight both the efficacy and efficiency of guided flows in model-based planning: a significant 10x speedup in offline RL with performance on par with diffusion models. This underscores the great potential of flow matching in extending the application of generative models to planning problems, especially those demanding enhanced computational efficiency, such as online planning.
\subsubsection*{Acknowledgments}
The authors thank Zihan Ding, Maryam Fazel-Zarandi, Brian Karrer, Maximilian Nickel, Mike Rabbat, Yuandong Tian, Amy Zhang, and Siyan Zhao for insightful discussions.

\bibliography{main}
\bibliographystyle{plainnat}

\newpage
\onecolumn
\appendix
\section{Proofs}



\subsection{Proof of Lemma 1}\label{a:lemma_vf_score}

\vfscore*

\textit{Proof (Lemma \ref{lem:vf_score}).} The Gaussian probability path $p_t(x|y)$ as in \eqref{e:p_t} is
    \begin{equation}
        p_t(x|y) = \int p_t(x|x_1) q(x_1|y)dx_1,
    \end{equation}
    where $p_t(x|x_1)=\gN(x|\alpha_tx_1,\sigma_t^2I)$. We express the score function as
    \begin{align}
        \nabla\log p_t(x|y) &= \frac{\nabla p_t(x|y)}{p_t(x|y)}\\
        &= \int \frac{\nabla p_t(x|x_1) q(x_1|y)}{p_t(x|y)}dx_1\\
        &= \int \nabla\log p_t(x|x_1)\frac{p_t(x|x_1) q(x_1|y)}{p_t(x|y)}dx_1.
    \end{align}
    The generating velocity field $u_t$ as in \eqref{e:u_t} is
    \begin{equation}
        u_t(x|y) = \int u_t(x|x_1) \frac{p_t(x|x_1)q(x_1|y)}{p_t(x|y)} dx_1,
    \end{equation}
    where $u_t(x|x_1)=\frac{\dot{\sigma}_t}{\sigma_t}(x-\alpha_tx_1) + \dot{\alpha}_tx_1$. Hence, by linearity of integrals it is enough to show that
    \begin{equation}
        u_t(x|x_1) = \frac{\dot{\alpha}_t}{\alpha_t}x + (\dot{\alpha}_t\sigma_t-\alpha_t \dot{\sigma}_t)\frac{\sigma_t}{\alpha_t}\nabla\log p_t(x|x_1).
    \end{equation}
    And indeed,
    \begin{align}
        u_t(x|x_1) &=\frac{\dot{\sigma}_t}{\sigma_t}(x-\alpha_tx_1) + \dot{\alpha}_tx_1\\
        &= \frac{\dot{\alpha}_t}{\alpha_t}x -\frac{\dot{\alpha}_t}{\alpha_t}x + \frac{\dot{\sigma}_t}{\sigma_t}(x-\alpha_tx_1) + \dot{\alpha}_tx_1\\
        & = \frac{\dot{\alpha}_t}{\alpha_t}x - (\dot{\alpha}_t\sigma_t-\alpha_t\dot{\sigma}_t)\frac{1}{\alpha_t\sigma_t}\parr{x - \alpha_tx_1}\\
        & = \frac{\dot{\alpha}_t}{\alpha_t}x + (\dot{\alpha}_t\sigma_t-\alpha_t\dot{\sigma}_t)\frac{\sigma_t}{\alpha_t}\nabla\log p_t(x|x_1),
    \end{align}
    where  in the last equality we used our assumption of Gaussian probability path that gives $\nabla\log p_t(x|x_1) = -\frac{1}{\sigma_t^2}(x-\alpha_tx_1)$. $\square$

\section{Probability Flow ODE for Scheduler $(\alpha_t, \sigma_t)$ }
\label{a:prob_flow_ode}
In this section we provide the velocity field used in CFG~\citep{ho2022classifier} for  approximate sampling from the conditional distribution $\qtilde(x|y) \propto q(x)^{1-\omega} q(x|y)^\omega$, $\omega\in \Real$, and show that it  coincides with our velocity field in \eqref{e:tilde_ut_score}. 

We assume the marginal probability paths (see \eqref{e:p_t}) $p_t(x)$ and $p_t(x|y)$ are defined with a scheduler $(\alpha_t,\sigma_t)$ and data distribution $q(x)$ and $q(x|y)$, respectively. CFG consider the probability path 
\begin{equation}
    \ptilde_t(x|y) = p_t(x)^{1-\omega} p_t(x|y)^\omega
\end{equation}
with the corresponding score function 
\begin{equation}
    \nabla \log \ptilde_t(x|y) = (1-\omega)\nabla p_t(x) + \omega \nabla p_t(x|y).
\end{equation}
Then the sampling is done with the Probability Flow ODE of diffusion models~\citep{song2020score}, 
    \begin{equation}\label{ae:dot_x_t}
        \dot{x}_t = f_t x_t -\frac{1}{2}g_t^2\nabla\log \ptilde_t(x_t|y),
    \end{equation}
    where $f_t = \frac{d\log\alpha_t}{dt}$, $g_t^2 = \frac{d\sigma_t^2}{dt} -2\frac{d\log\alpha_t}{dt}\sigma_t$~\citep{kingma2023variational,salimans2022progressive}.  Lastly,
    \begin{equation}
        \frac{d\log\alpha_t}{dt} = \frac{\dot{\alpha}_t}{\alpha_t}=a_t,\quad -\frac{1}{2}\frac{d\sigma_t^2}{dt} +\frac{d\log\alpha_t}{dt}\sigma_t =(\dot{\alpha}_t\sigma_t-\alpha_t\dot{\sigma}_t)\frac{\sigma_t}{\alpha_t}=b_t, 
    \end{equation}
    plugging this in \eqref{ae:dot_x_t} is an ODE with a velocity field that coincides with the velocity field in \eqref{e:tilde_ut_score}.

\section{Flow Matching Sampling with Guidance for Offline RL}
Comparing with standard generative modeling, the sequence model trained for RL needs to condition on the current state $s_t$, see Section~\ref{sec:rl}. Therefore, the sampling process is slightly different from Algorithm~\ref{algo:sampling_general}, as we need to zero out the vector fields corresponding to $s_t$, as shown in Algorithm~\ref{algo:sampling_rl}. Sampling for the goal-conditioned model can be done similarly.

\begin{algorithm}[H]
\DontPrintSemicolon
\small
\caption{Flow Matching Sampling with Guidance for Offline RL}
\label{algo:sampling_rl}
\textbf{Input:} 
initial state $s_0$, target return $R$,
guidance parameter $\omega$,
standard deviation of the starting distribution $\sigma$,
number of ODE steps $\node$\;
Sample $ x_0 \sim \N(0, \sigma^2 I)$ \;
$h \leftarrow 1/\node$
{\color{Green}\algorithmiccomment{step size}} \;
\For{$t = 0, h, \ldots, 1 - h$}{
    $x_t[0] \leftarrow s_0$ {\color{Green}\algorithmiccomment{fix the known token}} \;
    $ \tilde{u}_t(\cdot) \leftarrow (1 - \omega) \ut_t(\cdot) + \omega \ut_t(\cdot|R)$  {\color{Green}\algorithmiccomment{compute the velocity field under guidance}}\;
    Define $\bar{u}_t(\cdot): \R^p \mapsto \R^p$ such that for any $x\in \R^p$,
    the first dimension of $\bar{u}_t(x)$ is equal to $0$, and the other dimensions
    are the same as $\tilde{u}_t(x)$ 
    {\color{Green}\algorithmiccomment{zero out the VF for the known token}}\;
    $x_{t+h} \leftarrow $ ODEStep$(\bar{u}_t, x_t)$\;
}
\textbf{Output:} $x_1$
\end{algorithm}

\section{Image Generation Experiment Details}
We train three models on ImageNet-64: DDPM (using noise prediction), FM-CS, and FM-OT. 

FM-CS and FM-OT models are trained with the loss function in~\eqref{e:cfm_loss}:
\begin{equation}
\E_{t, b, q(x_1, y),p(x_0) } \norm{ \ut_t(x_t|(1-b)\cdot y + b \cdot \varnothing) - \dot{x}_t }^2,
\end{equation}
where $t$ is sampled uniformly in $[0,1]$, 
$b \sim \text{Bernoulli}(p_\text{uncond})$ is used to indicate whether we will use null condition,
$x_0$ is the noise, 
$x_1$ and $y$ are sampled from the true data distribution, and
$x_t = \alpha_t x_1 + \sigma_t x_0$, $\dot{x}_t=u_t(x_t|x_1)=\dot{\alpha}_t x_1 + \dot{\sigma}_t x_0$.
The noise scheduler of FM-CS is the cosine scheduler~\cite{albergo2022building}:
\begin{equation}
    \alpha_t = \sin \frac{\pi}{2} t, \qquad \sigma_t = \cos  \frac{\pi}{2} t,
\end{equation}
and the noise scheduler of FM-OT~\cite{lipman2022flow} is 
\begin{equation}
    \alpha_t = t, \qquad \sigma_t = 1 - t.
\end{equation}

DDPM models are trained with noise prediction loss as derived in \citet{ho2020denoising} and \cite{song2020score}:
\begin{equation}
    \E_{t, b, q(x_1, y), p(x_0) } \norm{ \eps^\theta_t(x_t|(1-b)\cdot y + b \cdot \varnothing) - x_0 }^2.
\end{equation}
We note that in our implementation, $t$ is sampled uniformly in $[0,1]$. We use the VP scheduler 
\begin{equation}
    \alpha_t = 1 - \zeta_{1-t}, \qquad \sigma_t = \sqrt{1 - \zeta^2_{1-t}}, \qquad \zeta_s = \exp{-\frac{1}{4} s^2(c_1-c_2) - \frac{1}{2}sc_2},
\end{equation}
with $c_1=20$ and $c_2 = 0.1$.

All three models have the same U-Net architecture adopted from \citet{dhariwal2021diffusion}, with hyperparameters listed below. For all the methods, we sweep the guidance weight across the range of 1.0 to 2.0 with a grid size of 0.05, and report the best results in Figure~\ref{fig:imagenet64_efficiency}. In particular, we have reported DDPM, DDIM and FM-CS using guidance weight 0.2, and FM-OT using guidance weight 0.15.
\begin{table}[H]
    \centering
    \begin{tabular}{l c}
        \toprule 
        hyperparameter & value \\
        \midrule
        channels & 196  \\
        depth &3 \\
        channels multiple & 1, 2, 3, 4\\
        heads & - \\
        heads channels & 64 \\
        attention resolution & 32, 16, 8\\
        dropout rate & 0.1\\
        batch size & 2048\\
        learning rate & 1e-4\\
        learning rate scheduler & constant\\
        iterations & $10^6$ \\
        $p_\text{uncond}$ & 0.2\\
         \bottomrule
    \end{tabular}
    \caption{Hyperparameters used to train diffusion models and flow models on the ImageNet-64 dataset.}
    \label{tbl:imagenet_hp}
\end{table}

\section{Offline RL Experiment Details}
\label{app:expr_details}
We summarize the architecture and other hyperparameters used for our experiments.
For all the experiments, we use our own PyTorch implementation that is heavily influenced by the following codebases: 

Decision Diffuser \hskip5pt \url{https://github.com/anuragajay/decision-diffuser} \\
Diffuser \hskip5pt \url{https://github.com/jannerm/diffuser/}\\

We train both guided flows and guided diffusion models for state sequences of length $H=64$.
The probability of null conditioning $p_\text{uncond}$ is set to $0.25$.
The batch size is $64$. We normalize the discounted RTG by a task-specific reward scale, which is $400$ for \hopper, $550$ for \walker and $1200$ for \cheetah. 
The final model parameter $\bar{\theta}$ we consider is an exponential moving average (EMA) of the obtained parameters over the course of training. For every $10$ iteration, we update 
$\bar{\theta} = \beta \bar{\theta} + (1- \beta) \theta$,
where the exponential decay parameter $\beta = 0.995$. 
We train the sequence model for $2 \times 10^6$ iterations, and checkpoint the EMA model every $200$k iteration.

\subsubsection*{Guided Flows}
We use a temporal U-net to model the velocity field $\ut$. 
It consists of 6 repeated residual blocks, where each block 
consists of 2 temporal convolutions followed by the group norm~\cite{wu2018group}
and a final Mish nonlinearity activation~\cite{mish2019self}.
The time $t$ is first trainsformed to its sinusoidal position encoding and projected to a latent space via a 2-layer MLP,
and the RTG $g(\vs)$ is transformed into its latent embedding via a 3-layer MLP.
The model is optimized by the Adam optimzier~\cite{kingma2014adam}. 
The learning rate is $2 \times 10^{-4}$ for \hopper-\medexpert, $3 \times 10^{-4}$ for \walker-\medreplay and $10^{-4}$ for
all the other datasets.

\subsubsection*{Guided Diffusion Models}
We use the cosine noise schedule proposed by \citet{nichol2021improved}.
We use a temporal U-net to model the noise $\eps_\theta$, with the same architecture used for guided flows.
The model is also optimized by the Adam optimzier, where the learning rate $2 \times 10^{-4}$ for all the datasets.

\subsubsection*{Inverse Dynamics Model}
The inverse dynamics model is modeled by an MLP with 2 hidden layers, 1024 hidden units per layer, and a $10\%$ dropout rate. We use the Adam optimizer with learning rate $10^{-4}$. We randomly sample 10\%
of offline trajectories as the validation set. We train the IDM for 100k iterations, and use the one that yields the best validation performance.

\section{Additional Experiments}
\label{app:additional_experiments}
\subsection{Flow Generated State Sequences}
\begin{figure}[H]
    \centering
    \begin{tabular}{c|c}
    \includegraphics[width=0.45\columnwidth]{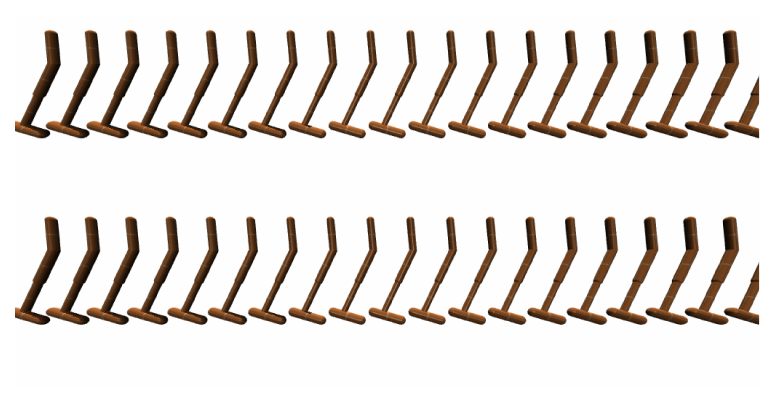} &
    \includegraphics[width=0.45\columnwidth]{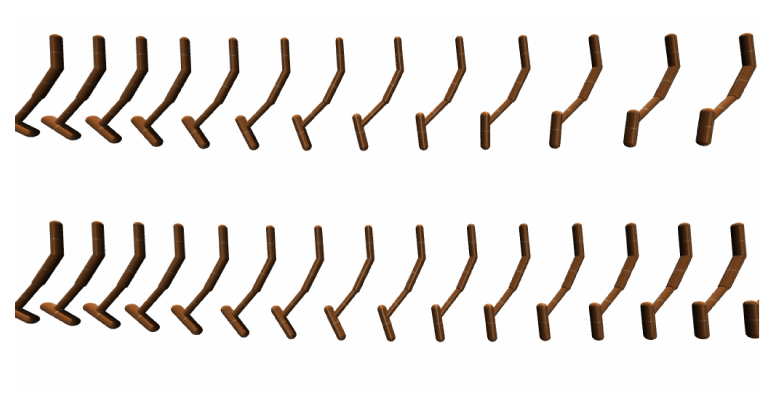} \\ \hline
    \includegraphics[width=0.45\columnwidth]{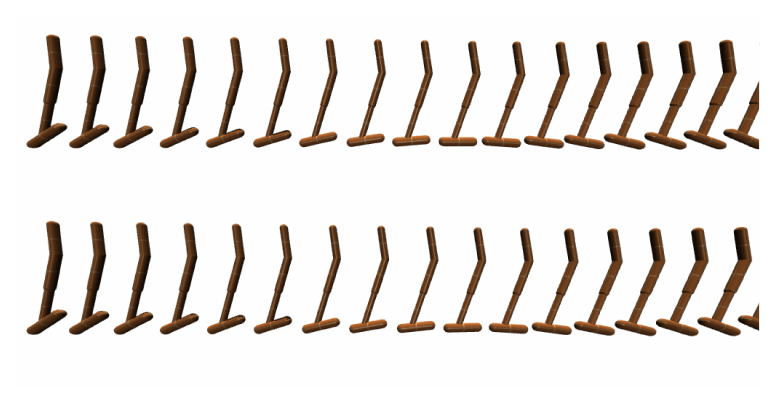} &
    \includegraphics[width=0.45\columnwidth]{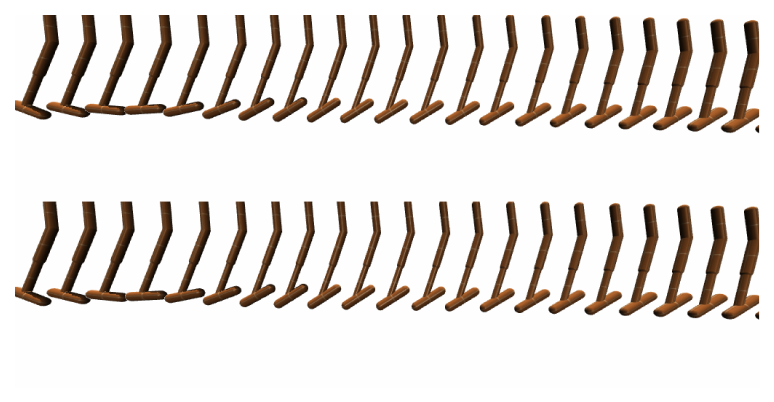} \\ \hline
    \includegraphics[width=0.45\columnwidth]{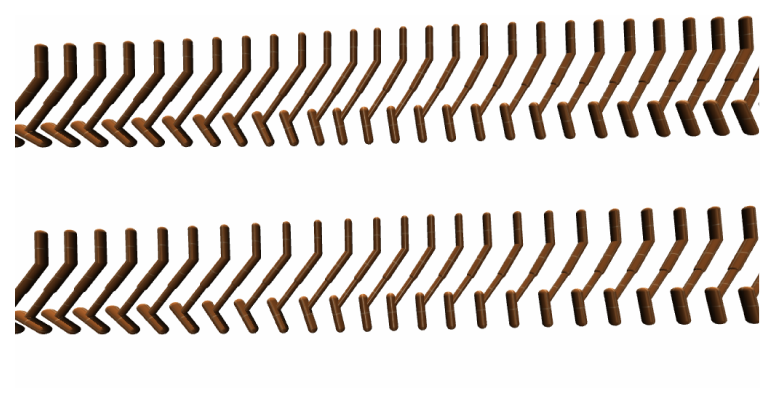} &
    \includegraphics[width=0.45\columnwidth]{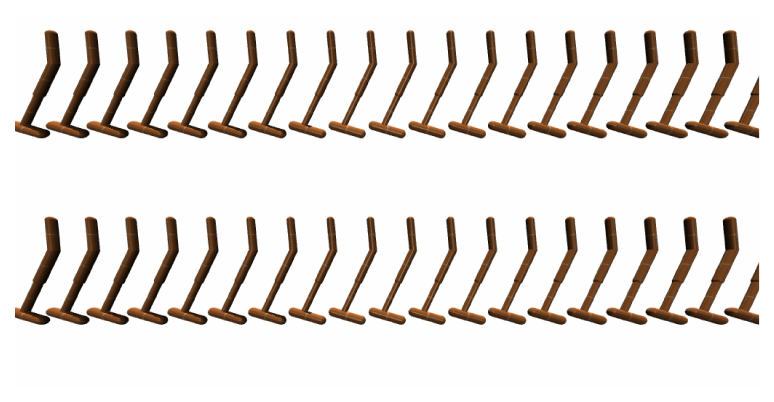} \\ \hline
    \includegraphics[width=0.45\columnwidth]{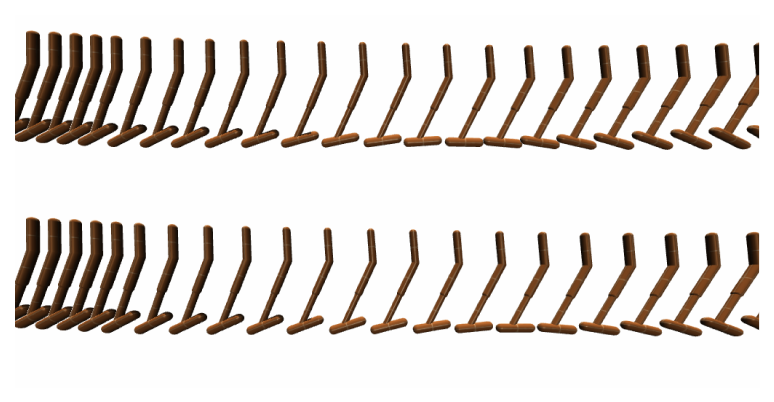} &
    \includegraphics[width=0.45\columnwidth]{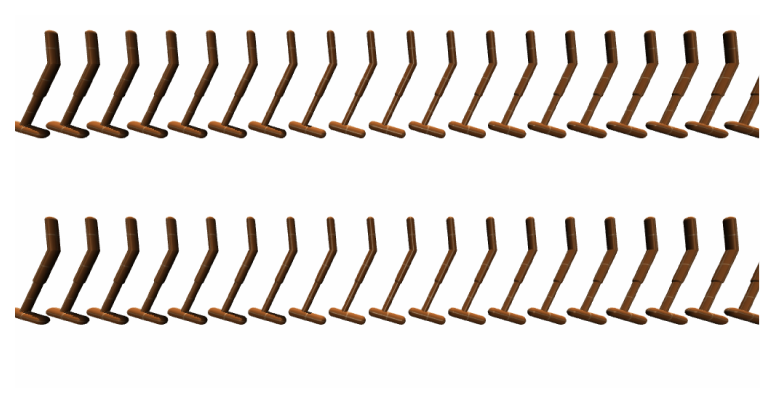}
    \end{tabular}
    \caption{Comparison of true state sequences and flow generated state sequences. In each panel, the top row plots
    a randomly sampled true state sequence from the \hopper-\medium dataset, and the bottom row plots the sequence sampled from the flow, conditioning on
    the first state and a large out-of-distribution RTG. The guidance weight is $3.0$.}
    \label{fig:hopper_trajs_more}
\end{figure}
\clearpage
\subsection{Computational Speed Comparison}
\begin{figure}[H]
    \centering
    \includegraphics[width=0.5\columnwidth]{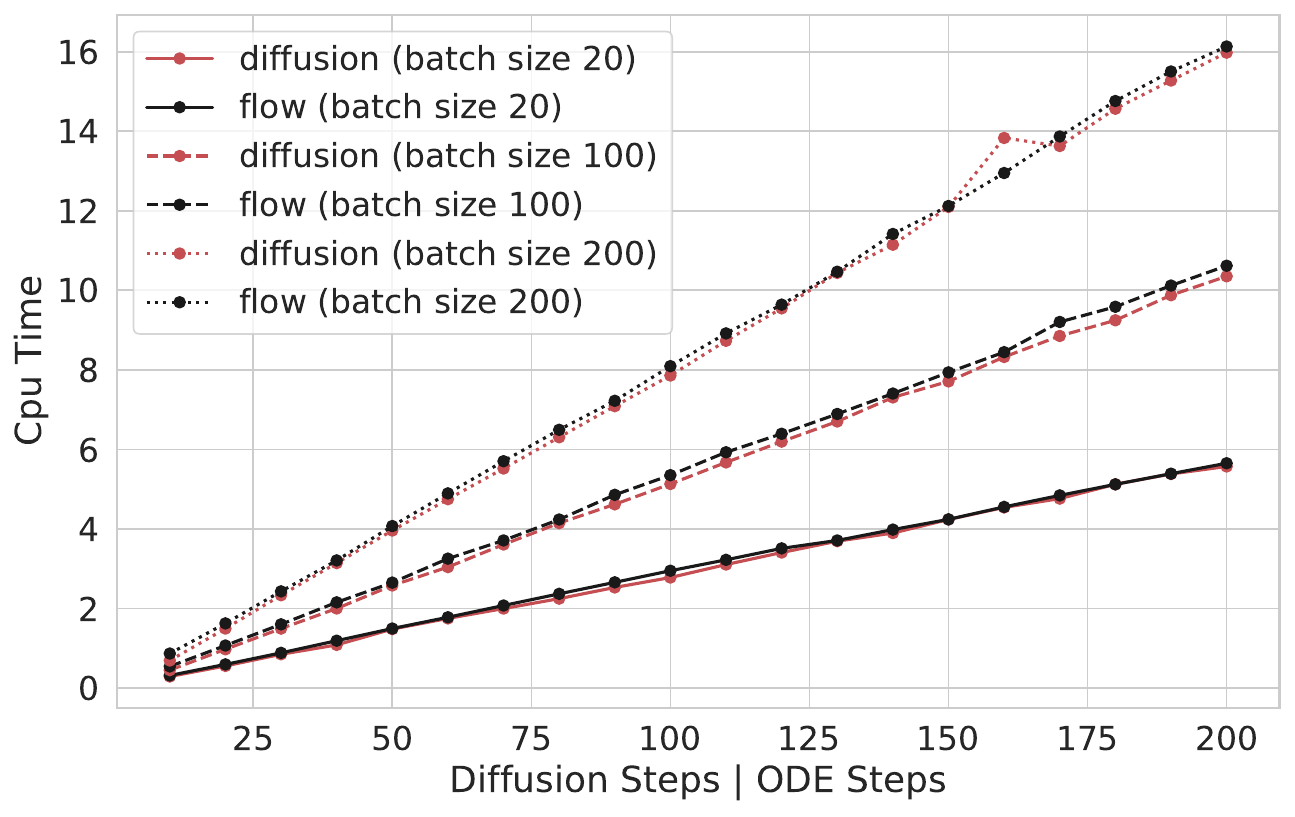}
    \caption{The CPU time of generating a batch of $704$-dimensional vectors (The subsequence length is $64$, and each state is of dimension $11$), using
    different numbers of internal steps. The time consumed by the diffusion model and the flow are roughly the same.}
    \label{fig:locomotion_time_comparison_all_batchsizes}
\end{figure}

\end{document}